\documentclass[11pt]{article}
\usepackage{tikzsymbols}
\usepackage{textcomp}

\usepackage{arabtex}
\usepackage{coling2020}
\usepackage{times}
\usepackage{url}
\usepackage{latexsym}
\usepackage[utf8]{inputenc}
\usepackage{hyperref}

\usepackage{utf8}
\usepackage{graphicx}
\usepackage{amsmath}
\usepackage[utf8]{inputenc}
\setcode{utf8}
\usepackage{multirow}
\usepackage[normalem]{ulem}
\useunder{\uline}{\ul}{}
\usepackage[titletoc,toc,title]{appendix}
\colingfinalcopy 

\title{Challenges in Translation of Emotions in Multilingual User-Generated Content: Twitter as a Case Study }

\author{
  Hadeel Saadany \thanks{*h.saadany@surrey.ac.uk} \\
  Centre for Translation Studies\\
  University of Surrey\\
  \\\And
  Constantin Or\u{a}san\\
  
  Centre for Translation Studies\\
  University of Surrey\\
  \\\And
  Roc\'io Caro Quintana\\  
  RGCL\\
  University of Wolverhampton\\
  
  \\\AND
  F\'elix do Carmo\\
  Centre for Translation Studies\\
  University of Surrey\\
  \\\And
  Leonardo Zilio\\
   Centre for Translation Studies\\
  University of Surrey\\
}

\date{}

\begin{document}
\maketitle
\begin{abstract}
  Although emotions are universal concepts, transferring the different shades of emotion from one language to another may not always be straightforward for human translators, let alone for machine translation systems. Moreover, the cognitive states are established by verbal explanations of experience which is shaped by both the verbal and cultural contexts. There are a number of verbal contexts where expression of emotions constitutes the pivotal component of the message. This is particularly true for User-Generated Content (UGC) which can be in the form of a review of a product or a service, a tweet, or a social media post. Recently, it has become common practice for multilingual websites such as Twitter to provide an automatic translation of UGC to reach out to their linguistically diverse users. In such scenarios, the process of translating the user's emotion is entirely automatic with no human intervention, neither for post-editing nor for accuracy checking. In this research, we assess whether automatic translation tools can be a successful real-life utility in transferring emotion in user-generated multilingual data such as tweets. We show that there are linguistic phenomena specific of Twitter data that pose a challenge in translation of emotions in different languages. We summarise these challenges in a list of linguistic features and show how frequent these features are in different language pairs. We also assess the capacity of commonly used methods for evaluating the performance of an MT system with respect to the preservation of emotion in the source text.
  
\end{abstract}

\section{Introduction}
\label{intro}

%
%
    %
    %
    %
    %
    %
    %

Despite the tremendous improvement in the quality of automatic translation as a result of the use of Neural Machine Translation (NMT) systems, NMT output can still contain errors. This is particularly noticeable with UGC such as tweets which do not follow common lexico-grammatical standards. Despite this limitation, NMT systems are commonly used in multilingual platforms such as Twitter to provide its users with an idea of global views or emotions towards current events or public figures. In such scenarios, the component of the tweet that conveys the emotion is often pivotal to the understanding of the tweet’s message. There have been different studies which explored how far sentiment information can be captured from the machine-translated text \cite{crosslingual,impact,howtrans,arabic_senti_analysis}. The objective of most research in this area, however, is from a sentiment classification perspective rather than a translation accuracy perspective. In these studies, the role of automatic translation of a language into English is simply to help in the sentiment classification of that language, since the available English sentiment classification systems can be directly applied to the target text \cite{trans_for_low,senti_after,withoutgood,evaluation}.

The research presented in this paper, however, evaluates the preservation of the sentiment message in translation not as a sentiment classification task but from a translation accuracy perspective. We assess how far automatic translation tools can be a successful real-life utility in transferring not only the sentiment polarity within tweets, but also fine-grained emotions such as anger and joy. The starting point is our work on translating Arabic UGC into English \cite{saadany2020great}. Analysis of sentiment mistranslations introduced by Google Translate when translating book reviews, revealed typical errors related to five linguistic phenomena: contronyms, diacritics, idiomatic expressions, dialectical code-switching, and negation. We have shown that the sentiment polarity of Arabic UGC is frequently flipped in the MT system output due to one of these five phenomena. In the current paper, we aim to investigate to what extent similar errors can be identified in the translation of tweets within various language pairs. We also go one step further by analysing the transfer of fine-grained emotions such as joy, anger, and fear by NMT systems. To achieve this, we carry out an analysis of datasets of tweets automatically translated into different language pairs. At the end of this analysis, we attempt to provide answers to the following questions:

\begin{enumerate}
\item Are there specific linguistic features of tweets that can lead to mistranslation of emotions? 
\item  Do these features affect different languages equally?
\item Can traditional automatic quality measures adequately evaluate the translation of sentiment?

\end{enumerate}

To answer the above research questions, this paper is divided as follows: section \ref{data} presents our data compiling and experiment for evaluating the translation of emotion in tweets by MT systems. Section \ref{analysis} analyses challenging features for the translation of emotions in multilingual tweets. It also provides a qualitative analysis of each feature based on its frequency in our compiled dataset and its prominence in each of the source language explored. In section \ref{measure}, we evaluate the efficacy of the MT automatic quality metrics in assessing the mistranlation of emotion within the multilingual UGC framework. Section \ref{limit} presents a conclusion on the our experiment, limitations of the present study and recommendations for future research work.

\section{Data Collection and Experiment Setup}
\label{data}
In order to check how far automatic translation captures the specific emotion in tweets, we
replicated a real-life scenario where MT systems are utilised spontaneously to translate the
content of tweets. Twitter currently supports built-in translations, so users can click on a
\textit{Translate Tweet} prompt visible directly under the tweet text to translate it. Twitter mentions
that it employs Google Translate for this service. To evaluate how far the MT system in this
scenario can serve as a real-life utility, we used Google Translate API to automatically
translate compiled multilingual Twitter datasets previously annotated for four emotions (joy,
fear, aggression, and anger). It is important to note that these four emotions were chosen as
representative of the common fine-grained sentiments expressed in tweets. The authors of
tweets are usually either happy, angry, or fearful of something or someone, and their anger
can either be aggressive or passive.
The datasets were collated from different emotion-detection and aggression-detection shared
tasks \cite{shared1,shared2,shared3,shared4}. The source datasets amounted to approximately 30,000
tweets in three languages: English, Arabic, and Spanish.

We created two datasets from this
source annotated data. The first dataset was created by translating the Arabic and Spanish
source datasets into English. The second dataset was created by translating part of the source
English dataset into Romanian, Arabic, Spanish and Portuguese. These datasets were used to
extract instances for our analysis.
The next stage in our experiment was to extract instances in which the MT system failed to
translate the emotion correctly. We call this failure “mistranslation of emotion” and it is
identified by the discrepancy between the annotations of emotion in the source dataset and the
emotions classified in the translated tweets. For example, if the original tweet is annotated as
conveying ‘anger’ but the classifier predicted ‘joy’ for the translation, this pair was
considered a potential mistranslation of emotion and was selected for analysis. 

To get the
classifications of emotions in the translated tweets, we built a classifier by fine-tuning a
Roberta XML model \cite{roberta} on the previously annotated 23,000 source English
tweets. This data was pre-processed by deletion of punctuation, non-alphanumeric symbols,
lemmatisation, and lower-casing. We also used the Demoji\footnote{\url{https://pypi.org/project/demoji/}} Python library to transfer the emojis
into their equivalent lexicon (e.g. \dSadey[1.5] is translated into ``dislike”). We tested the accuracy of
the classifier against a test set of these tweets, and its accuracy was 92\%. The English
classifier was used to predict the emotion of the Google API output for the translation of the
Arabic and Spanish dataset into English and the back translation of the English tweets
translated into the other languages. The classifier’s predicted emotion was compared to the
gold standard emotion of the source text, instances of discrepancy were extracted as potential
mistranslations of emotions.

\section{Analysis of Challenging Features for the Translation of Emotions in Tweets}
\label{analysis}
\begin{figure}[t]
\centering
\includegraphics[scale=0.50,trim={0.2cm .2cm .3cm .1cm},clip]{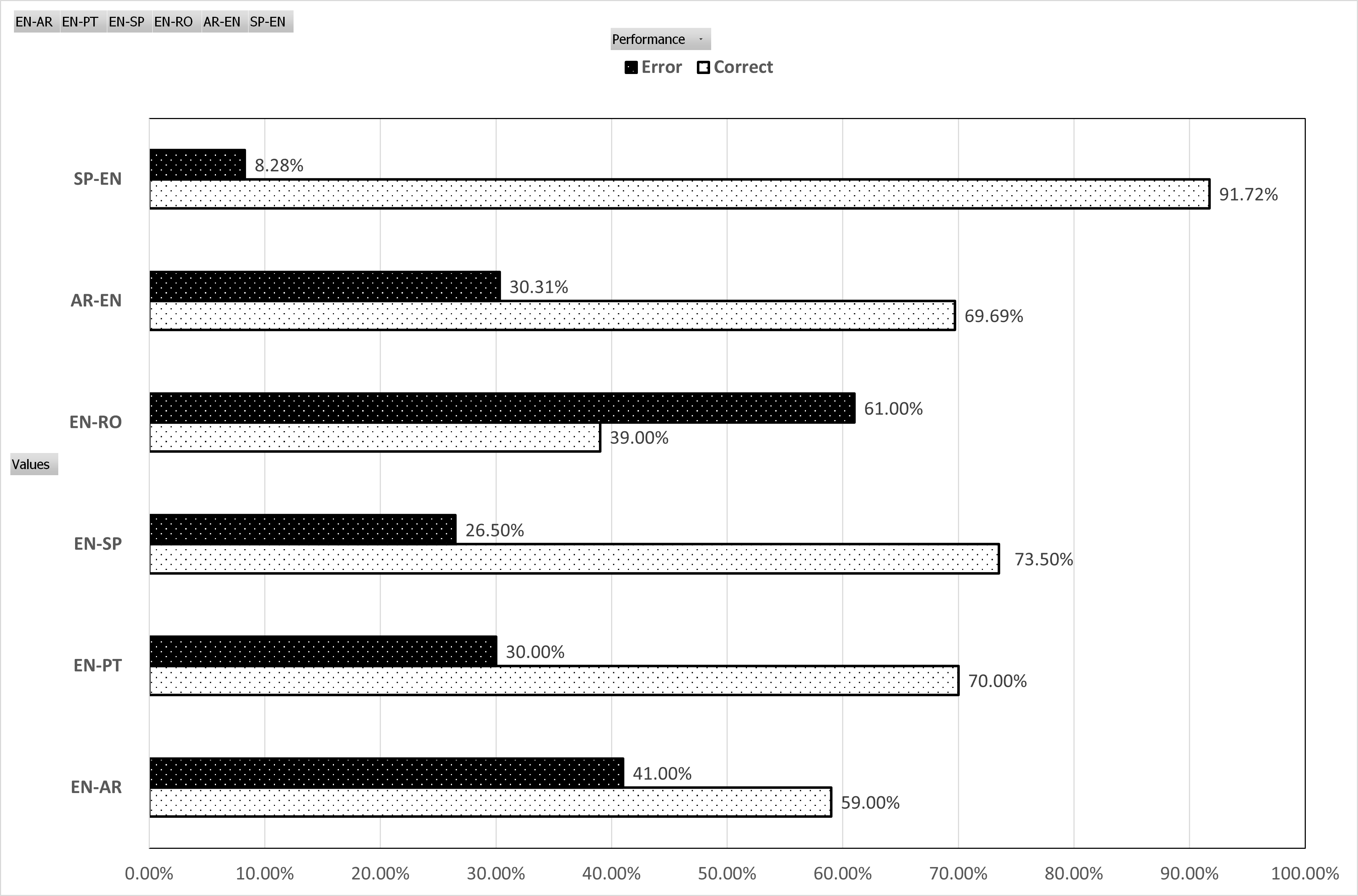}
\caption{Frequency of Mistranslation of Emotion in the Analysed Dataset}
\label{Freq}
\end{figure}

To check the reasons for discrepancy between the predicted emotion and the emotion of the source text, a team of MT researchers, native speakers of the respective languages, conducted a manual analysis on samples of the extracted potential mistranslations. The extracted samples for English to other languages amounted to 1600 tweets divided equally among the four target languages, and from the opposite direction, with English as a target language, it amounted to $\approx3000$ tweets divided between Arabic and Spanish as a source language. The disagreements in the dataset due to mistranslation of emotions are presented in Figure \ref{Freq}. Spanish has clearly fewer cases of discrepant emotions in tweets, both when these are translated into English ($\approx8$\%) and when they are translations from English ($\approx27$\%). Target languages like Romanian (61\%) and Arabic (41\%) show a much higher percentage of tweets with mistranslated emotions. It is obvious from the analysed sample that some languages are more privileged than others in the real-life scenario we replicate for our experiment.

Next, we decided to look in detail to linguistic features in the source tweets that could help explain the mistranslation of emotion. Based on an initial analysis of the sample dataset and previous studies of the language of tweets, we selected six features that represent special challenges in transferring emotions by the MT engine: hashtags, slang, non-standard orthography, idiomatic expressions, polysemy, and grammar (negation structures). The following sections demonstrate the effect of these features on the translation of emotion with illustrative examples . Table \ref{tab1} presents a summary of our findings, which are discussed next.

\begin{table}[]
\centering
\resizebox{\textwidth}{!}{%
\begin{tabular}{|l|l|l|l|l|l|l|}
\hline
\textbf{Language Pair} & \textbf{Hashtags} & \textbf{Slang} & \textbf{Polysemy} & \textbf{Idiomatic   Expressions} & \textbf{Grammar} & \textbf{Orthography} \\ \hline
\textbf{EN-SP} & \textbf{44\%} & 14\%         & 7.9\%          & 6.3\%         & 12.6\%          & 14\%            \\ \hline
\textbf{EN-PT} & 41.6\%         & 16.6         & 2.7\%           & 8.3\%         & \textbf{13.8\%} & 16.6\%          \\ \hline
\textbf{EN-AR} & 25.6\%         & 20.7\%        & \textbf{24.3\%} & \textbf{12\%} & 6\%             & 11\%            \\ \hline
\textbf{EN-RO} & 24.6\%         & 26\%          & 18.6\%          & \textbf{12\%} & 6\%             & 12.6\%          \\ \hline
\textbf{AR-EN} &               & \textbf{60\%} & 11\%            & 7.9\%         & 6.7\%           & 13.9\%          \\ \hline
\textbf{SP-EN} &               & 32.5\%        & 16\%            & 16.5\%        & 12\%            & \textbf{22.6\%} \\ \hline

\end{tabular}%
}
\caption{Frequency of Language Features per Language Pair}
\label{tab1}
\end{table}

\subsection{Hashtags}
Emotions in tweets are expressed in a special style in line with Twitter's orthographic limitations and peculiarities. Thus, for example, authors of tweets frequently express their emotion as a trailing hashtag or a hashtagged non sequitur to a neutral or an ironic statement. The emotion of the tweets in such cases is retrieved solely from the hashtag. Our analysis has shown that this unique style of emotion transfer constitutes a challenge to the MT system. When the hashtags expressing emotion are either untranslated or mistranslated, the emotion expressed in the message is completely distorted. For example, the fear emotion in the English tweet ``Just waved daughter and her friend off to school, \#terrifying!" is entirely missed in the Arabic translation ``\< لقد لوحت ابنتها وصديقتها الصغيرة للتو  إلى المدرسة > \#terrifying!" as the  hashtag that carries the main emotional content is not translated. Moreover, the hashtagged word in tweets is often written in non-standard orthography which causes the MT to output the hashtagged word as is without translation. For example, the anger emotion against customer service in the tweet ``I asked for my parcel to be delivered to a pickup store not my address \#poorcustomerservice" is missed in the Romanian translation ``Am cerut livrarea coletului meu la un magazin de preluare, nu adresa mea \#poorcustomerservice" as the hashtagged word is not translated. The MT treats such hashtags as out-of-vocabulary words and hence misses the affective message. The distortion of emotion is also caused by a wrong translation of the hashtagged word. The anger emotion of the English tweet “CNN's Wolf Blitzer calls you an American astronaut and you don't correct him \#disappointed” is completely lost in the Spanish translation as the hashtag is mistranslated to ``diseñado" meaning `designed' instead of disappointed. The Spanish translation carries a neutral emotion. 

\subsection{Slang and Dialectical Expressions} 

Research studies have shown that slang and dialectical expressions present several challenges to MT in general \cite{zbib2012machine}. Tweets are characterised by the wealth of slang expressions and the code-switching between different dialects of one language based on the authors’ demographics. It was observed from the manual analysis of the sample data that this stylistic quirk often distorts the translation of emotion in the source text. For example, the Spanish tweet ``Ni en pedo, bueno en pedo si" is mistranslated in English as ``not even fart good fart yes". The correct translation of the expression ``ni en pedo" is `no way'. The source tweet expressed a humorous comment which should read ``No way. Well, yes way". In this example, the MT online engine provides an incomprehensible output due to a mistranslation of the dialectical version of the Spanish expression ``ni en pedo" used mainly in Argentina, and therefore the emotion of the source text is completely lost. Similarly, the MT system fails to detect the aggression in the English tweet ``The iconic nigger tweet" when it is translated to Romanian as ``tweet-ul iconic negru" (The iconic black tweet). The slang expression in the source tweet (nigger) carries the aggressive tone and hence the neutral translation (black) misses the aggressive emotion.
The amount of distortion of the affective messages due to a mistranslation of slang or dialectical expressions varies from one language to another. It was observed that Arabic dialectical expressions posed a significant challenge to the MT system as it caused a distortion of emotion in 60\% of the Arabic tweets in the second dataset (see Table \ref{tab1}). For example, commenting on an event in the Middle East, a tweeter expresses joy ``\<ايه كمية الانشكاح ديه>" (What all this amount of happiness!). The MT system gives the exact opposite emotion ``What all this amount of anger!". This owes to the fact that the dialectical expression ``\<الانشكاح>" (happiness) is mistranslated as ``anger". The dialectical tweets were mostly mistranslated in aggressive Arabic tweets. For example, bullying a female football player, a tweet says ``\<جايه بفستانها وتستلم جايزة افضل لاعبة خربوا الكورة الحريم>" (She is coming with a dress to receive the best player prize, women ruined football). The tweet is written in a Gulf dialect that was mistranslated by the MT engine as “come to her dress and receive the prize for the best player who ruined the harem?”. The MT output not only misses the aggressive tone but also lacks semantic and grammatical coherence.

\subsection{Non-standard Orthography}

With its 280-character limit, Twitter users often resort to creative abbreviations and unconventional orthography. Moreover, linguists have observed that to encourage speed and immediacy of understanding, Twitter users type in the same way they speak \cite{theguardian_2010}. The manual analysis has shown that this specific linguistic phenomenon is a major culprit in a wrong transfer of the emotion within different language pairs. For example, the MT output of the English tweet ``watching sad bts video bc im sad. iwannacryy" renders an incomprehensible affect message in Portuguese (assistindo ao vídeo do sad bts bc im sad. iwannacryy). The reason is that the microblogging limitation causes the author of the tweet to use a creative word shortening by eliminating spaces ``iwannacryy" as well as by texting in acronyms (``bc" meaning ``because", ``im" meaning ``I am"). The affective message is missed in the Portuguese translation as all these emotional nuanced orthographic forms remain untranslated. Another complication is that tweeters are more apt to use expressive lengthening to communicate strong emotion. These non-standard emotional expressions are usually treated as out-of-vocabulary by MT systems with all the language pairs the research team analysed. For instance, the anger in the Spanish tweet ``Por que sos re chantaaaa" (Why are you such a liar?) is not transferred by the MT translation ``Why are you chantaaaa" as the Spanish word ``chanta" (liar) passes for out-of-vocabulary lexicon because of elongation. 

\subsection{Idiomatic Expressions}

One of the challenging issues in the field of translation is the process of translating the different shades of meaning conveyed by an idiom \cite{almubarak}. The reason is that translating idioms usually involves meta-linguistic information such as cultural and social norms. Because of its informal nature, conversational idioms are used extensively in tweets. The manual analysis has shown that a large number of idioms were literally translated, which did not only affect the sentiment preservation of the source text, but often produced nonsensical target text. For example, the Arabic tweet expressing happiness towards one particular public figure ``\<والله دمه خفيف>" has the idiomatic expression ``\<دمه خفيف>", meaning ``funny". The tweet should read `By God, he is so funny', but the MT output gives a literal translation, ``By God, his blood is so light" which was predicted as having an `anger' sentiment by the classifier. The same problem also exists in language pairs with English as a source language. For example, an `angry' tweet commenting on the recent American presidential elections ``We have to keep u in line" has the idiomatic expressions ``keep in line" meaning to discipline uncontrolled behaviour. This idiom was literally translated in Arabic as ``\<في الطابور>" (stay in the queue) and in Spanish as “mantenerte en línea” (stay fit). The literal translation of the idiom in the two language pairs flips the emotion from anger to a neutral sentiment.

\subsection{Polysemy}

MT research has shown that polysemous words pose a challenge to MT systems when the contextual information is not clearly determined \cite{polysemy}. Due to its micro-blogging nature, polysemous words in tweets are usually lacking context, which adds to their ambiguity. The manual analysis of the translated data has revealed that this linguistic phenomenon distorts the tweeter’s emotional message. One example is the aggressive English tweet ``the girl sitting in front of me is chewing her gum like a cow; I’m ready to snap". The word snap here has the informal meaning of ``burst in anger". The Romanian translation by the MT system, however, reflects a joy emotion as it gives the other meaning of snap ``take pictures". Hence the MT Romanian output reads ``the girl in front of me chews her gum like a cow; I'm ready to take pictures". Another more extreme example appears with the Arabic to English pair. Commenting on a Middle East political crisis, an aggressive tweeter threatens two Gulf countries ``\<جايكم الدور خلينا نربيكم في اليمن ونربي قطر>" (Your turn will come, Yemen and Qatar, we will teach you a lesson). The aggressive threat is lost in the MT output ``Come on let’s educate Yemen and Qatar". This is due to a mistranslation of the polysemous word ``\<نربي>" which could either positively mean ``educate" or to negatively mean ``teach a lesson" by inflicting punishment.

\subsection{Grammar (Negation)}

The analysis has shown that the distortion of the source emotion was also associated with a wrong translation of a negation marker between different language pairs. For example, negation was found to cause a problem when the source text is in dialectical Arabic. The lexico-grammatical realisation of negation differs between the Standard and dialectical Arabic as well as between its different dialects. Arabic dialects often treat negative particles as clitics, and hence a letter is added to the stem of the word to change it to negative \cite{emad}. The MT engine frequently either missed the Arabic dialectical negation and hence flips the phrase to an opposite sentiment pole or mistranslates the negated phrase altogether. For example, commenting on a terrorist attack, a tweeter angrily states ``\<لا سامح الله من افتخر بقنبلة دمرت مئات المنازل>" (May God not forgive (punish) the one who is proud of a bomb destroying hundreds of homes). The negation is missed and hence the online translation tool output reads ``May God forgive that one who is proud of a bomb destroying hundreds of homes". The emotion of the tweeter in the Arabic translation is flipped from anger to sympathy towards a terrorist attack. If automatic translation is used to spot potential terrorist trends on social media platforms, this type of error would affect the accuracy of the algorithm and may bring dangerous consequences to users. The analysis has also shown that missing negation structures in the English tweets also distorts the emotion. For example, the fear emotion in the English tweet ``A trip to the dentist never gets easier" is flipped to joy in the Portuguese translation because of a wrong translation of the negative structure. The MT output in Portuguese is ``Uma ida ao dentista nunca foi tão fácil" meaning `A trip to the dentist has never been easier'.  

\section{Measuring the transfer of sentiment}
\label{measure}

From the analysis of these language features, it can be observed that using automatic translation tools for translating emotion in multilingual UGC such as tweets involves several linguistic challenges. Despite these challenges, NMT systems such as Google Translate are extensively utilised by social media platforms without human post-editing. In the research environment, the reliability of MT systems is commonly determined by automatic quality metrics that are domain agnostic as they evaluate the translation accuracy regardless of the type of source text. In this section we touch upon the issue of evaluating the translation of sentiment-oriented text by MT engines. We will show that the standard quality metrics for MT evaluation fall short of their goal in capturing mistranslations of emotions in special types of text such as Twitter data.

The de facto standard for MT performance evaluation is the BLEU score with its different variations \cite{papineni2002bleu}. BLEU gives equal penalty weight to inaccurate translation of n-grams, which may lead to performance overestimation (or underestimation). For example, the “joy” emotion in a tweet from our Arabic dataset (``What is this amount of happiness, I don’t understand!") is flipped to anger by Twitter’s Google Translate tool which outputs ``What is this amount of anger, I don’t understand!". Despite the distortion of the affective message, BLEU only mildly penalises the swapping of the two opposite emotive nouns `happiness' and `terrible' and this translation receives a BLEU score of 0.76. The reason is that BLEU gauges the performance of an MT model by an indiscriminate n-gram matching, regardless of the semantic weight of each word. By human standards, the MT performance in such cases is highly over-estimated.

There have been numerous efforts to address the common pitfalls of n-gram-based metrics by incorporating semantic and contextual features in metrics such as METEOR \cite{banerjee2005meteor}. Although the introduction of more semantically oriented metrics showed a better correlation with human judgement, the estimation of sentiment preservation in UGC has not yet been addressed. When it comes to evaluating an MT system performance in transferring emotion, even the semantically oriented automatic metrics do not give a penalty to a mistranslated sentiment proportional to the distortion it afflicts on the source message. For example, the negation in the Arabic tweet ``May God do not forgive those who put you in power" is missed in the MT output, ``May God forgive the one who put you in power". The emotion is flipped from anger to joy. Despite the distortion of the emotion, the mistranslation receives a METEOR score 0.61. 

For this last analysis, we selected an evaluation dataset consisting of 100 tweets extracted from the dataset that was classified by the research team as having a mistranslation of emotion. The tweets in this dataset were chosen in a way where the main error in the translation is the distortion of emotion due to one of the six linguistic features discussed in the previous sections. The evaluation dataset was translated by native speakers in the research team. We calculated both BLEU and METEOR segment scores of the machine-translated tweets against the human references. The average BLEU and METEOR scores per segment is approximately 0.60 and 0.45 respectively, with a standard deviation of 0.2 for BLEU and 0.1 for METEOR. These high values show that the replacement of a sentiment by its opposite in a translated tweet, as we had in this evaluation dataset, is not captured by lexical similarity metrics. This calls for an improvement of evaluation metrics, introducing mechanisms more sensitive to distortion of emotion in UGC data such as tweets.

\section{Limitations and Future Work}
\label{limit}

In this research, we evaluated the capacity of an MT online system in translating the features of tweets that convey emotions. Our analysis highlighted a number of linguistic challenges involved in translating emotion between different languages. Future research may point to other types of features that may prove significant for other language pairs. Moreover, using NMT tools in translating raw text such as sentiment-oriented tweets may prove counterproductive, as it may provide users with an affective message that is different or even opposite to the source emotion. Future research may test whether preprocessing of raw UGC such as Twitter data can improve the MT performance in transferring the correct emotion. 

We also touched upon the reliability of automatic quality measures for evaluating MT systems performance in transferring emotion. We have shown that the standard evaluation measures were not able to give a penalty proportional to the incorrect translation of emotion in a sample dataset of mistranslated tweets. We believe that evaluating the performance of MT systems in translating sentiment-oriented text is an under-recognised problem in MT research. Future work should address the possibility of introducing a sentiment measure to reflect how far the MT system transfers the correct affective message in the source text.







\bibliographystyle{coling}
\bibliography{coling2020}

\end{document}